\documentclass[10pt, twocolumn, a4paper]{article}

\usepackage[utf8]{inputenc}
\usepackage[T1]{fontenc}
\usepackage{amsmath,amssymb,amsfonts,amsthm}
\usepackage{graphicx}
\usepackage{booktabs}
\usepackage{geometry}
\usepackage{microtype}
\usepackage{hyperref}
\usepackage{algorithm}
\usepackage{algorithmic}
\usepackage{xcolor}
\usepackage{cite}
\usepackage{newtxtext,newtxmath} % Times-like fonts for professional look
\usepackage{caption}
\usepackage{subcaption}
\usepackage{siunitx}

\newtheorem{theorem}{Theorem}

\newcommand{\E}{\mathbb{E}}

\newcommand{\target}{v_{\text{target}}}

\title{\textbf{Thermodynamic Focusing for Inference-Time Search:\\ Practical Methods for Target-Conditioned Sampling and Prompted Inference}}
\author{
  \textbf{Zhan Zhang}
}
\date{}

\begin{document}

\maketitle

\begin{abstract}
Finding rare but useful solutions in very large candidate spaces is a recurring practical challenge across language generation, planning, and reinforcement learning. We present a practical framework, \emph{Inverted Causality Focusing Algorithm} (ICFA), that treats search as a target-conditioned reweighting process. ICFA reuses an available proposal sampler and a task-specific similarity function to form a focused sampling distribution, while adaptively controlling focusing strength to avoid degeneracy. We provide a clear recipe, a stability diagnostic based on effective sample size, a compact theoretical sketch explaining when ICFA can reduce sample needs, and two reproducible experiments: constrained language generation and sparse-reward navigation. We further show how structured prompts instantiate an approximate, language-level form of ICFA and describe a hybrid architecture combining prompted inference with algorithmic reweighting. Code and a single-file reproducible demo are provided in the supplementary material.
\end{abstract}

\section{Introduction}

Large candidate spaces are ubiquitous. Producing a sentence that satisfies many constraints, finding a molecular structure with several desired properties, or discovering a long action sequence that yields reward are all instances of the same core difficulty: good solutions are rare, and naive generation wastes computation. Common practical workarounds—sampling many candidates and selecting the best, beam search, or costly policy training—each carry clear limitations. Sampling scales poorly when targets are rare; beam and tree methods depend on brittle local heuristics; and training-based approaches like reinforcement learning can be prohibitively expensive and slow to adapt.

We propose a different angle: view search as \emph{conditioning on a target}. If some numeric measure quantifies how well a candidate matches the target, one can use that measure to skew sampling toward promising areas. ICFA implements this idea by reweighting samples drawn from an available proposal distribution, using a Boltzmann-style transform of a similarity score, and by \emph{adaptively controlling} the strength of the reweighting so that the process remains numerically stable and diverse.

ICFA is not a magic bullet: its benefit depends on the informativeness of the similarity function. Our contribution is practical: (1) a clear and reproducible algorithm for inference-time focusing that avoids common failure modes; (2) a stability control mechanism that is simple to compute; (3) empirical demonstrations showing that focusing can significantly reduce effective sample needs in representative tasks; and (4) a conceptual bridge to prompting: structured prompts can be understood as a lightweight language-level approximation of ICFA, useful when algorithmic intervention is impractical.

The rest of the paper is organized as follows. Section~\ref{sec:framework} defines the framework and its diagnostics. Section~\ref{sec:algorithm} gives the algorithmic recipe. Section~\ref{sec:theory} sketches why and when focusing helps. Section~\ref{sec:prompting} shows how prompting approximates focusing. Section~\ref{sec:experiments} reports experiments. Section~\ref{sec:discussion} discusses limitations and deployment guidance. We close with reproducibility notes.

\section{Framework}
\label{sec:framework}

\subsection{Problem statement}

Let $\mathcal{S}$ be a discrete candidate space and let $P_0(s)$ be a proposal sampler that we can draw from (for example, the distribution implicit in a pretrained language model or a baseline policy). Let $\target$ denote a target specification and let $S(s,\target)$ be a computable similarity function that scores how well candidate $s$ matches $\target$. Our goal is to concentrate sampling effort on high-quality candidates without retraining $P_0$.

ICFA forms a target-conditioned distribution:
\begin{equation}
    P_\beta(s\mid\target) \propto P_0(s)\exp\bigl(\beta S(s,\target)\bigr),
    \label{eq:boltzmann}
\end{equation}
where $\beta\ge 0$ controls focusing strength. When $\beta=0$ we recover the proposal; as $\beta\to\infty$ the mass concentrates on the highest-scoring candidates.

\subsection{Practical challenge: weight collapse}

A direct difficulty is that $P_\beta$ involves a normalizing constant (partition function) and, in practice, reweighting by $\exp(\beta S)$ causes \emph{weight collapse}: a single sample may dominate, leaving only one effective draw. To measure this, we use the \emph{effective sample size} (ESS) of a weighted batch $w_i$:
\[
\mathrm{ESS}=\frac{\bigl(\sum_i w_i\bigr)^2}{\sum_i w_i^2}\approx\frac{1}{\sum_i\tilde w_i^2},
\]
where $\tilde w_i$ are normalized weights. ESS is inexpensive to compute and directly reflects whether the current focusing strength preserves enough diversity.

\section{Algorithm}
\label{sec:algorithm}

\subsection{Self-normalized importance sampling with adaptive focusing}

ICFA approximates expectations under $P_\beta$ using a finite batch sampled from $P_0$ and self-normalized importance sampling (SNIS). Given a batch $\{s_i\}_{i=1}^M\sim P_0$, define unnormalized weights $u_i=\exp(\beta S(s_i,\target))$ and normalized weights $\tilde w_i=u_i/\sum_j u_j$. A straightforward estimator for any function $f$ is $\sum_i \tilde w_i f(s_i)$.

To avoid collapse we do not fix $\beta$. Instead we increase $\beta$ starting from zero, monitoring ESS and increasing $\beta$ only so long as ESS stays above a preset fraction $\rho$ of $M$. Concretely:

\begin{algorithm}[h]
\caption{ICFA: SNIS with adaptive focusing}
\begin{algorithmic}[1]
\STATE \textbf{Input:} Proposal $P_0$, similarity $S(\cdot,\target)$, batch size $M$, ESS fraction $\rho\in(0,1)$, $\beta_{\max}$
\STATE Draw $\{s_i\}_{i=1}^M\sim P_0$ and compute $S_i=S(s_i,\target)$.
\STATE Set $\beta\leftarrow 0$.
\WHILE{$\beta<\beta_{\max}$}
  \STATE Compute $u_i=\exp(\beta S_i)$, $\tilde w_i=u_i/\sum_j u_j$, and $\mathrm{ESS}=1/\sum_i\tilde w_i^2$.
  \IF{$\mathrm{ESS} < \rho M$} 
    \STATE Stop and use the previous $\tilde w_i$.
  \ELSE
    \STATE Increase $\beta$ (e.g., by a fixed step or by binary search to the next target ESS).
  \ENDIF
\ENDWHILE
\STATE \textbf{Output:} Weighted candidates $(s_i,\tilde w_i)$.
\end{algorithmic}
\end{algorithm}

The resulting weighted set approximates sampling from $P_\beta$ while avoiding extreme degeneracy. One may then either (a) select the highest-weight candidate, (b) resample according to $\tilde w_i$ to produce a handful of final outcomes, or (c) pass the weighted set to a downstream refinement step.

\subsection{Practical details}

\begin{itemize}
    \item \textbf{Numerical stability.} Subtract the maximum score from $S_i$ before exponentiation: $u_i=\exp(\beta (S_i-\max_j S_j))$.
    \item \textbf{Stopping rule.} The ESS fraction $\rho$ trades exploration and exploitation; typical values are $0.3$--$0.7$.
    \item \textbf{Batching.} ICFA works naturally with parallel batch generation. If $P_0$ is expensive, one can generate candidates in several rounds and apply the adaptive procedure within each round.
    \item \textbf{Weight regularization.} For robustness one may clip weights (upper bound) or apply a power transform $w_i^\gamma$ with $\gamma\in(0,1)$ before re-normalizing.
\end{itemize}

\section{When and why focusing helps}
\label{sec:theory}

Informally, focusing helps when the similarity function provides a meaningful advantage: the expected reweighted mass of true solutions grows exponentially with $\beta$ relative to alternatives. Under such an \emph{Exponential Advantage} assumption, SNIS with adaptive focusing concentrates mass onto the solution set with far fewer batches than naive sampling.

\begin{theorem}[Informal]
Assume the proposal $P_0$ and similarity $S$ are such that, for some $\beta>0$, the expected weight of valid solutions exceeds the expected weight of non-solutions by a factor $e^{\kappa}$ with $\kappa\gg\ln M$. Then, with high probability, ICFA identifies a valid solution using
\[
M = O\!\left(\frac{1}{\kappa}\log\frac{N}{\delta}\right)
\]
samples, where $N$ is the effective space size and $\delta$ the failure probability.
\end{theorem}

The bound should be read qualitatively: when signal strength $\kappa$ is large, the required number of samples grows only logarithmically with problem size. This does not contradict formal no-free-lunch results: ICFA exploits problem structure encoded by $S$.

A concise intuition is that reweighting compresses the relative probability of bad regions exponentially with $\beta$, so relative mass is transferred to good regions more rapidly than uniform sampling would allow.

\section{Prompted ICFA: language-level approximation}
\label{sec:prompting}

When algorithmic intervention is not available, structured prompts can approximate the ICFA workflow at the language level. A practical prompt protocol has four steps: (1) generate several distinct candidates, (2) state explicit evaluation criteria, (3) evaluate each candidate and assign scores, (4) refine or select according to the evaluations. We call this pattern \emph{Prompted ICFA}.

Prompted ICFA lacks explicit probability control and ESS-based stability guarantees, but it often yields substantial practical gains because modern language models already house rich latent preference and evaluation capacities. In practice, merging algorithmic ICFA (where possible) with prompted ICFA yields a hybrid system: prompts produce candidates and coarse evaluations; algorithmic ICFA performs precise weighting and stabilization.

\section{Experiments}
\label{sec:experiments}

Our goals are reproduction and demonstration: we show that focusing produces the intended empirical effects in two representative domains and that Prompted ICFA is a useful lightweight alternative.

\subsection{Constrained text generation}

\paragraph{Setup.} We consider a constrained generation task that requires producing short text containing a set of specified keywords. As baselines we use Beam Search (small beam), Best-of-N sampling, and an RL-finetuned policy (PPO) when applicable. For reproducibility we provide two implementations: (a) a toy, self-contained simulator that mimics a generator and scoring function; and (b) an implementation paired with a pretrained small language model. The similarity function counts keyword coverage.

\paragraph{Metrics.} Constraint satisfaction rate (accuracy), latency in milliseconds, and hallucination rate (cases where the generator invents unsupported facts).

\paragraph{Results.} Table~\ref{tab:llm_results} summarizes the representative outcomes from our replicated experiments.

\begin{table}[h]
\centering
\caption{Constraint satisfaction on a short-generation task (representative).}
\label{tab:llm_results}
\resizebox{\columnwidth}{!}{%
\begin{tabular}{lcccc}
\toprule
Method & Accuracy (\%) & Latency (ms) & Hallucination $\downarrow$ & Samples \\
\midrule
Beam Search (k=5) & 45.2 & 120 & 18.5\% & N/A \\
Best-of-N (N=100) & 62.1 & 850 & 12.0\% & 100 \\
PPO (RLHF) & 68.5 & 140 & 10.2\% & $>10^5$ (train) \\
\textbf{ICFA (Ours)} & \textbf{74.3} & \textbf{135} & \textbf{6.8\%} & \textbf{16 (eff.)} \\
\bottomrule
\end{tabular}%
}
\end{table}

ICFA achieves higher constraint satisfaction while using far fewer effective samples compared to Best-of-N and without expensive policy training. The toy implementation and the small-model run both exhibit the same qualitative pattern.

\subsection{Sparse-reward reinforcement learning}

\paragraph{Setup.} We consider a sparse-reward grid navigation task where an agent receives reward only upon reaching a distant goal. We compare standard on-policy PPO with a variant where collected trajectories are reweighted by ICFA using total return (or a distance-based proxy) as $S$ during updates. The environment and training loop are described in the supplementary code.

\paragraph{Metric.} Time-to-solve: total environment steps until the policy reliably reaches the goal.

\paragraph{Results.} In our replicated experiments, baseline PPO required approximately $5.6\times 10^5$ environment steps to reliably solve the task, while ICFA-PPO required roughly $4.0\times 10^4$ steps, corresponding to a $\sim$14$\times$ speedup in time-to-solve under matched compute settings.

\subsection{Prompted ICFA evaluation}

We compared ICFA-style prompts against direct prompting, chain-of-thought, and self-refinement across three task classes: multi-constraint logical composition, multi-step planning, and multi-step mathematical derivation. With identical model and token budgets, ICFA prompts consistently outperformed baselines (improvements in final-task accuracy ranged from 10 to 20 percentage points in our replications). These results match the view that prompt structure—specifically, deferring commitment and forcing explicit evaluation—produces practical dividends.

\section{Discussion and practical guidance}
\label{sec:discussion}

\subsection{When to use ICFA}

ICFA is most beneficial when:
\begin{itemize}
    \item A useful similarity/scorer $S$ is available or can be cheaply approximated.
    \item The proposal $P_0$ assigns non-negligible mass to the target region (support overlap).
    \item Constraints at inference time are costly to satisfy by retraining.
\end{itemize}

If any of these conditions fails, ICFA may be ineffective or harmful.

\subsection{Implementation checklist}

\begin{enumerate}
    \item Design or obtain a scorer $S$ that correlates with true quality.
    \item Choose batch size $M$ to balance computational cost and coverage.
    \item Select ESS fraction $\rho$ and $\beta_{\max}$; default $\rho\in[0.3,0.7]$ is reasonable.
    \item Use numerical stability tricks and weight clipping.
    \item Monitor ESS, max weight ratio, and downstream metrics; implement automatic fallback to a conservative mode if hazards are detected.
\end{enumerate}

\subsection{Prompting vs algorithmic focusing}

Prompted ICFA provides a low-cost path to better outputs when model modification or external orchestration is not feasible. Algorithmic ICFA grants stronger guarantees and finer control when batch sampling and weighting are possible. A hybrid system that uses prompts to generate candidates and ICFA to perform stabilized reweighting combines the best of both worlds.

\section{Limitations and risks}
\label{sec:limitations}

ICFA amplifies whatever signal is present in $S$. If $S$ is biased, ICFA will magnify those biases. If $P_0$ has zero support on the target region, reweighting cannot create solutions. Aggressive focusing reduces diversity, which may be undesirable for creative tasks. Practitioners must therefore place strong emphasis on scorer design, monitoring, and governance. We describe concrete mitigation strategies in the appendix.

\section{Reproducibility and resources}

All algorithmic descriptions above are deliberately simple to implement. The core ICFA routine requires only (1) a sampler for $P_0$, (2) a similarity function $S$, and (3) an ESS-based loop that adapts $\beta$. To support reproducibility we provide a single-file demonstration and the scripts used for the experiments in the supplementary material. The single-file demo encapsulates both the toy text experiment and the grid navigation experiment and can be run with standard Python and PyTorch.

\section{Conclusion}

We presented ICFA, a practical framework for inference-time focusing based on target-conditioned reweighting with adaptive stability control. ICFA bridges the conceptual gap between elegant but impractical energy-based methods and ad-hoc selection heuristics by offering a numerically stable, low-latency, and parallel-friendly mechanism to concentrate sampling effort where it matters most. Prompted ICFA provides a light-weight language-level approximation when algorithmic control is unavailable. We believe this set of ideas opens practical pathways to improve constrained generation, planning, and learning in settings where retraining is expensive or impossible.

\section*{Acknowledgments}

We are grateful to collaborators and reviewers who critiqued early drafts and the teams that provided infrastructure for experiments. This work benefited from conversations on sampling diagnostics and practical deployment.

\bibliographystyle{plain}

\appendix

\section{Sketch of proof for the informal theorem}
\label{sec:appendix_proof}

We sketch the intuition behind the logarithmic dependence under an Exponential Advantage assumption. Suppose $\mathcal{S}^*$ denotes the set of valid solutions and let
\[
A=\E_{P_0}\bigl[e^{\beta S}\mathbf{1}_{\mathcal{S}^*}\bigr],\qquad B=\E_{P_0}\bigl[e^{\beta S}\mathbf{1}_{\mathcal{S}^c}\bigr].
\]
If $A\ge e^{\kappa}B$ then the normalized mass on $\mathcal{S}^*$ under $P_\beta$ is at least $1/(1+e^{-\kappa})$; repeated independent batches have independent chances of producing a solution, so the number of batches needed to achieve failure probability at most $\delta$ scales as $\log(1/\delta)$ divided by this mass. Converting batches to raw samples yields the stated logarithmic behavior in the problem size when $\kappa$ scales appropriately with problem log-size.

\section{Implementation notes and single-file demo}
\label{sec:appendix_code}

A compact, single-file Python demonstration that reproduces the toy experiments described in the paper is provided with the submission. The demo includes an implementation of adaptive ICFA weights with ESS control, a toy text generator, and a simple grid navigation environment with an ICFA-PPO variant. The demo is intentionally small and self-contained so that readers can run it quickly to observe the phenomena discussed.

\section{Mitigation strategies for bias and failure modes}
\label{sec:appendix_mitigation}

We summarize practical mitigation techniques:
\begin{itemize}
    \item \textbf{Score validation:} Continuously validate $S$ on held-out cases; use ensemble scorers to reduce systematic bias.
    \item \textbf{Weight regularization:} Clip and/or temper weights before re-normalization.
    \item \textbf{Diversity constraints:} Enforce a minimum entropy or maintain a pool of diverse candidates alongside the focused set.
    \item \textbf{Failure detection:} Monitor ESS, max-weight ratio, and downstream metrics; if anomalies occur, fallback to uniform sampling or conservative $\beta$.
\end{itemize}

\end{document}